\documentclass[lettersize,journal]{IEEEtran}
\usepackage{amsmath,amsfonts}
\usepackage{algorithmic}
\usepackage{algorithm}
\usepackage{array}
\usepackage{booktabs}
\usepackage{subfigure}
\usepackage{textcomp}
\usepackage{multirow}
\usepackage{makecell}
\usepackage{stfloats}
\usepackage{url}
\usepackage{verbatim}
\usepackage{graphicx}
\usepackage{cite}
\usepackage[colorlinks=true, allcolors=blue]{hyperref}
\begin{document}


\title{Text-guided Eyeglasses Manipulation with Spatial Constraints}

\author{Jiacheng Wang, Ping Liu*, Jingen Liu, Wei Xu*,
\thanks{J. Wang and W. Xu are with the Hubei Key Laboratory of Smart Internet Technology, School of Electronic Information and Communications, Huazhong University of Science and Technology, Wuhan 430074,
China, e-mail: jiacheng@hust.edu.cn, xuwei@hust.edu.cn.}
\thanks{P. Liu is with the Center for Frontier AI Research (CFAR), Research Agency for Science, Technology and Research (A*STAR), Singapore 138634, e-mail: pino.pingliu@gmail.com.}

\thanks{J. Liu is with Disney Streaming Advanced Research, USA, email: jingen.liu@gmail.com.}

\thanks{* means co-corresponding author.}}

\markboth{Journal of \LaTeX\ Class Files}%
{Shell \MakeLowercase{\textit{et al.}}: A Sample Article Using IEEEtran.cls for IEEE Journals}


\maketitle

\begin{abstract}

Virtual try-on of eyeglasses involves placing eyeglasses of different shapes and styles onto a face image without physically trying them on. 
While existing methods have shown impressive results, the variety of eyeglasses styles is limited and the interactions are not always intuitive or efficient. 
To address these limitations, we propose a Text-guided Eyeglasses Manipulation method that allows for control of the eyeglasses shape and style based on a binary mask and text, respectively. 
Specifically, we introduce a mask encoder to extract mask conditions and a modulation module that enables simultaneous injection of text and mask conditions. 
This design allows for fine-grained control of the eyeglasses’ appearance based on both textual descriptions and spatial constraints.
Our approach includes a disentangled mapper and a decoupling strategy that preserves irrelevant areas, resulting in better local editing. 
We employ a two-stage training scheme to handle the different convergence speeds of the various modality conditions, successfully controlling both the shape and style of eyeglasses. 
Extensive comparison experiments and ablation analyses demonstrate the effectiveness of our approach in achieving diverse eyeglasses styles while preserving irrelevant areas.
\end{abstract}


\section{Introduction}

\IEEEPARstart{V}{irtual} try-on technology allows individuals to virtually add fashion items to their personal images, facilitating the assessment of suitability and appeal. 
Specifically, virtual try-on for eyeglasses enables users to experiment with a variety of styles without physical presence, \textit{e.g.,} sunglasses, metal glasses, and eyeglasses of different colors.
This approach is highly efficient and convenient, as it eliminates the need for individuals to physically try on numerous eyeglasses to determine the appropriate style and shape. 
Moreover, virtual try-on technology for eyeglasses has gained popularity due to the emergence of short videos, as it enables users to apply various styles to recorded videos for enhanced visual effects, commonly referred to as ``eyeglasses special effects" in different applications.

Several approaches have been proposed to improve eyeglasses virtual try-on technology \cite{3DVTON1, 3DVTON2, 3DVTON3, 3DVTON4, 3DVTON5, hisd, interfacegan, ganspace}. 
These approaches can be broadly classified into two categories: 3D-based and 2D-based methods. 
3D-based methods \cite{3DVTON1, 3DVTON2, 3DVTON3, 3DVTON4, 3DVTON5} employ 3D eyeglasses models to align the eyeglasses with the face. 
In contrast, 2D-based methods \cite{interfacegan, ganspace, sefa, attgan, hisd} directly edit input images using generative adversarial networks (GANs) \cite{gan} to produce the desired eyeglasses, which generally rely on the learned latent space to perform eyeglasses manipulation.
However, a limitation of these methods is that the introduction of new eyeglasses styles (\textit{e.g.,} sunglasses, metal glasses, and eyeglasses of different colors) typically requires either building new 3D eyeglasses models \cite{3DVTON1, 3DVTON2} for 3D-based methods or recalculation of editing directions in the latent space \cite{interfacegan, ganspace} for 2D-based methods. 
This can be inefficient and inconvenient to scale.

One alternative solution to this issue is to leverage text input as a means of conveniently scaling diverse eyeglasses styles. 
Vision-Language models, such as Contrastive Language-Image Pre-training (CLIP)  \cite{clip}, have enabled various methods for arbitrary text-guided image manipulation \cite{styleclip, ppe, cf-clip, no_token, feat, deltaedit, fusedream, stylegan-nada}. 
StyleCLIP \cite{styleclip}, in particular, is a pioneering work in this area, leveraging the powerful text-to-image alignment capabilities of CLIP to manipulate original images based on text descriptions. 
{Subsequently, many methods\cite{ppe, no_token, feat, cf-clip, deltaedit, fusedream, stylegan-nada} have improved StyleCLIP to achieve better image editing results. However, these methods are primarily designed for general face manipulation and are rarely used for diverse eyeglasses manipulation. To ensure an intuitive and convenient eyeglasses virtual try-on process, it is imperative to conduct further research on controlling eyeglass styles through text prompts.}


\begin{figure}[!t]
    \centering
    \includegraphics[width=0.5\textwidth]{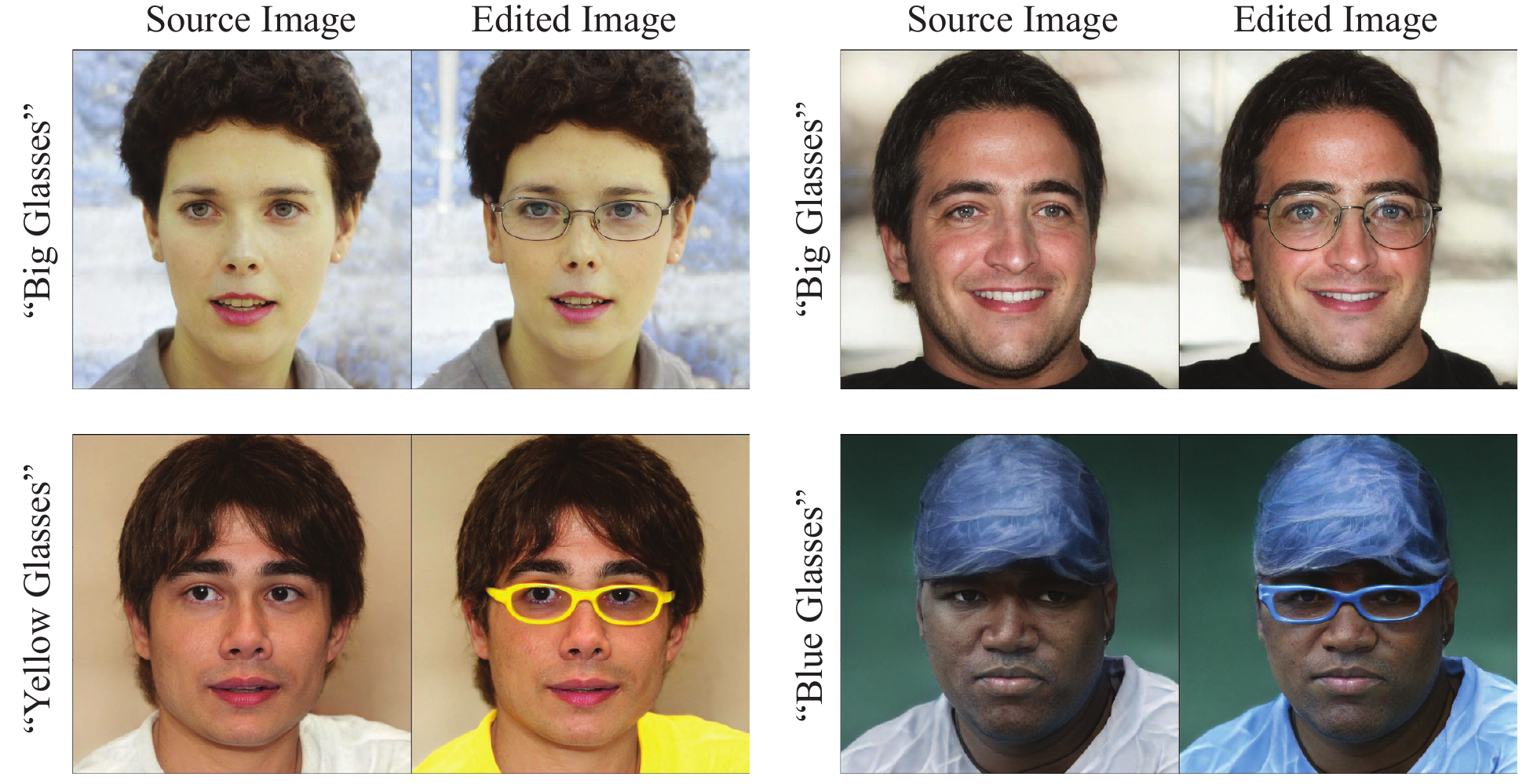}
    \caption{{Issues existing on text-guided virtual try-on for eyeglasses.}
    The first row depicts that eyeglasses in the editing results are of different sizes based on the same text prompt of ``big glasses". 
    The second row shows an entangled phenomenon that irrelevant areas are significantly modified, especially in the cloth region.}
    \label{fig:introduction_issues}
\end{figure}

{As we directly applying these methods to eyeglass manipulation tasks, they suffers from several drawbacks, as illustrated by some results from StyleCLIP\cite{styleclip} shown in Fig. \ref{fig:introduction_issues} (refer to Fig. \ref{fig:qualitative_results} for more comparisons).}
Firstly, it can be difficult to express the degree of editing precisely, particularly for the spatial configuration (\textit{e.g.}, sizes, shapes) of eyeglasses, which is critical for eyeglasses virtual try-on. 
For instance, using the same text prompt of ``big glasses" may produce different editing results with eyeglasses of varying sizes, indicating an ambiguity challenge that size words such as ``small", ``medium", and ``large" may not be consistent with the absolute pixel sizes of items in images\cite{vl-checklist}.
Secondly, modifying the eyeglasses region conditioned on text prompts often leads to noticeable changes in unrelated areas, such as the clothing region, which is a common entangled phenomenon in many existing methods\cite{styleclip, ppe, cf-clip}. 
In this manner, it can be challenging to succinctly modify specific areas to accommodate the given condition while maintaining the consistency of unrelated regions, which requires specific design.

To address the ambiguity challenge and disentanglement challenge, we designed a text-guided eyeglass manipulation method with spatial constraints. 
Inspired by HairCLIP\cite{hairclip}, we harness the remarkable expression capacity of textual descriptions to manipulate diverse styles of eyeglasses using a single model. 
However, different from HairCLIP, our approach integrates spatial constraints through the use of eyeglasses masks to handle the ambiguity challenge.
Specifically, the mask is automatically generated by an off-the-shelf face parser on the FFHQ\cite{stylegan} and CelebA-HQ\cite{celeba-hq} datasets, which are then used to construct aligned data pairs based on the original face pose.
To address the disentanglement challenge, we view the editing process as a two-step procedure, with one step concentrating on concisely altering eyeglasses and the other on maintaining the consistency of unrelated areas.
As a result, we specifically design the following components and training strategy to solve the eyeglass editing challenges.

Firstly, we introduce a mask encoder comprising multiple convolutional blocks to extract mask conditions, which are then employed to control the spatial configuration of eyeglasses for more precise manipulation. 
Simultaneously, we design a modulation module to allow for the simultaneous injection of text and mask conditions. 
Utilizing the binary eyeglasses mask and straightforward text prompts, we can effortlessly regulate the shape and style of eyeglasses within a single model. 

Secondly, to address the entangled issue, we propose GlassMapper, which consists of an editing mapper and a disentangled mapper to enable more distinct and disentangled editing directions. 
The editing mapper is responsible for controlling the eyeglasses style, while the disentangled mapper preserves irrelevant areas. 
We achieve this by employing a simple yet effective decoupling strategy, which involves truncating the gradient flow from the disentangled mapper to the editing mapper.

Thirdly, we adopt a two-stage training scheme to address the varying convergence speeds of different modalities. 
This enables us to sequentially equip the GlassMapper with the ability to modify eyeglasses, resulting in a stable training process.
After completing the two-stage training, we can simultaneously manipulate the style and shape of eyeglasses based on different modality conditions.

In summary, our contributions are as follows:
\begin{itemize}
    \item We propose a Text-guided Eyeglasses Manipulation method to achieve diverse and flexible eyeglasses virtual try-on. 
    Our method controls eyeglasses shape and style based on a simple binary eyeglasses mask and text description, respectively. 
    This approach accommodates a wide range of eyeglasses shapes and common styles within a single model, offering a more convenient and intuitive mode of interaction.
    \item To support multiple modalities of information in a single model, we propose a new modulation module that combines the binary eyeglasses mask and natural language descriptions. Furthermore, we employ a two-stage training scheme to stabilize the training process, sequentially equipping our method with the ability to modify eyeglasses. 
    \item {To better preserve irrelevant areas, we introduce a disentangled mapper and a simple decoupling strategy, resulting in better local editing. 
    This allows for more accurate and precise control over the eyeglasses style and shape, while also maintaining the integrity of the surrounding areas in the image.}
    \item Extensive quantitative and qualitative experiments on the CelebA-HQ dataset \cite{celeba-hq} are conducted to demonstrate the superiority of our approach. 
    On average, we outperform the state of the arts \cite{ppe} by 9.21\%, 29.32\%, and 11.13\% on \textit{Structure Similarity Index Measure} (SSIM)\cite{ssim}, \textit{Peak Signal to Noise Ratio} (PSNR)\cite{psnr}, and \textit{Identity Discrepancy Scores} (IDS), respectively.
\end{itemize}


\section{Related work}

\subsection{Eyeglasses Virtual Try-on}

The goal of eyeglasses virtual try-on is to add eyeglasses with a specific style to the face of an image or video, which is opposite to eyeglasses removal\cite{ergan, byeglassesgan}. 
This can be accomplished through two main categories of methods: 3D-based eyeglasses virtual try-on and 2D-based eyeglasses virtual try-on.
3D-based eyeglasses virtual try-on methods \cite{3DVTON1, 3DVTON2, 3DVTON3, 3DVTON4, 3DVTON5} typically add eyeglasses to a face image by aligning the 3D models of eyeglasses and face. 
For example, Milanova et al. \cite{3DVTON1} proposed a markerless eyeglasses virtual try-on system that overlays the 3D eyeglasses model over the face image according to the estimated head pose. 
Similarly, Feng et al. \cite{3DVTON2} achieved 3D eyeglasses virtual try-on by combining 3D face reconstruction and pose estimation techniques. 
While there are many real-time eyeglasses virtual try-on systems based on 3D models, each change of eyeglasses style requires a new 3D eyeglasses model, which can be difficult to collect and scale.

2D-based eyeglasses virtual try-on methods \cite{interfacegan, ganspace, sefa, attgan, hisd} directly edit the given images to obtain the desired eyeglasses. 
Some methods \cite{interfacegan, ganspace, sefa} use pre-trained GANs, specifically StyleGAN \cite{stylegan, stylegan2}, and learn an editing direction in their latent space to control the generated eyeglasses style. 
Others \cite{attgan, hisd} mainly train an encoder and a generator from scratch, where the encoder extracts the latent representation of the original image and the generator achieves arbitrary eyeglasses style manipulation based on different attribute constraints. 
However, the variations of generated eyeglasses in these methods are limited, \textit{e.g.}, only black plastic eyeglasses or sunglasses can be added in most cases. 
To handle this limitation, a concurrent work GlassGAN \cite{glassesgan} focuses on multi-style eyeglasses virtual try-on using pre-trained StyleGAN \cite{stylegan, stylegan2}. 
They explore the editing directions targeted toward eyeglasses in an unsupervised setting, \textit{i.e.}, solving an eigen-problem. 
However, they need to subjectively assign human-interpretable attributes to the explored editing directions, such as size, position, squareness, roundness, cat-eye appearance, and thickness, which are limited to express the spatial consistency of generated eyeglasses. 

\begin{figure*}[!t]
    \centering
    \includegraphics[width=\textwidth]{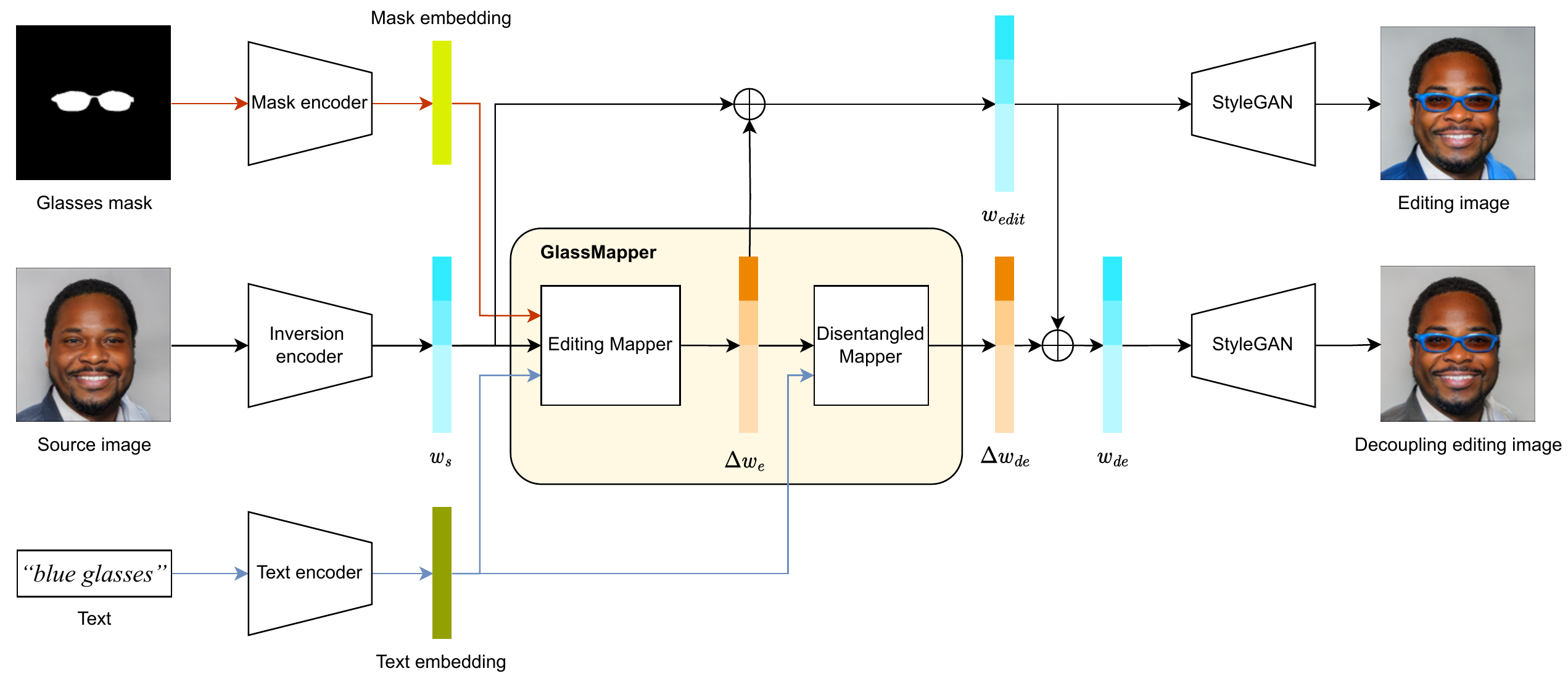}
    \caption{An overview of our approach.
    Our approach takes the source latent code $w_s$, mask embedding $e_m$, and text embedding $e_t$ as input through a StyleGAN inversion encoder, mask encoder, and CLIP text encoder, respectively.
    Later, the GlassMapper predict an editing latent code $w_{edit}$ and a decoupling editing latent code $w_{de}$, which are then applied to the StyleGAN generator to generate the editing image $I_{edit}$ and the decoupling editing image $I_{de}$.
    Specifically, $I_{edit}$ achieves the desired manipulation but unavoidably suffers significant modification in unrelated regions. In contrast, $I_{de}$ inherits the impressive manipulation capability while preserving irrelevant areas well.
    The details of GlassMapper are provided in Section \ref{framework}.}
    \label{fig: overview framework}
\end{figure*}

\subsection{Text-guided Image Manipulation}

{With the development of generative models, there has been a significant amount of research on facial image synthesis and manipulation. 
The objective of facial image synthesis is to produce realistic and high-fidelity portraits. 
Unconditional methods\cite{stylegan, stylegan2} emphasize the diversity and quality of the resulting images, whereas conditional methods\cite{anyface, tedigan, sketch_face_1, sketch_face_2, sketch_face_3, sean, semanctic_synthesis} prioritize the ability to control the generated images by utilizing textual descriptions\cite{anyface, tedigan}, sketches\cite{sketch_face_1, sketch_face_2, sketch_face_3}, or semantic maps\cite{sean, semanctic_synthesis}.}
Moreover, facial image manipulation\cite{sketch_guided, text-guided, appearance-transfer, fine-grained, image_guided} endeavors to alter facial attributes in accordance with user predilections, including images\cite{appearance-transfer, image_guided}, sketches\cite{sketch_guided}, and textual descriptions\cite{text-guided}.

In pursuit of a more convenient and intuitive means of 
manipulation interaction, text-guided image manipulation aims to modify source images according to simple text descriptions, while preserving irrelevant areas.
With the powerful capabilities of CLIP\cite{clip}, StyleCLIP \cite{styleclip} proposed a CLIP loss to make the editing results consistent with the given text. 
Later, to better preserve irrelevant areas and learn a more disentangled editing direction, several works \cite{ppe, no_token} proposed disentangled loss functions, and \cite{feat} learned an attention mask to limit editing areas. 
Given that the original CLIP loss proposed in StyleCLIP is vulnerable to adversarial samples, \cite{cf-clip, fusedream, stylegan-nada} modified the CLIP loss using data augmentation or contrastive learning to improve robustness.

Several works \cite{anyface, bridge_clip, deltaedit, maniclip} achieved arbitrary text-guided image manipulation without inference-time optimization or restricting to single attribute editing. 
Specifically, \cite{anyface, bridge_clip, deltaedit} focused on bridging the latent space of StyleGAN \cite{stylegan, stylegan2} and CLIP \cite{clip}, while \cite{maniclip} injected text conditions into an attention decoder to obtain corresponding editing directions. 
With the development of diffusion models, several works \cite{diffusionclip, ldedit, blenddiffusion} combined CLIP with diffusion models \cite{ddpm, ddim, ldm} to handle the limitations of GAN inversion \cite{diffusionclip} and extend the image manipulation domain \cite{ldedit, blenddiffusion}. 
HairCLIP \cite{hairclip} is another recent work that focuses on the diversity of specific attributes, specifically hair attributes, which achieves control of $44$ hairstyles and $12$ hair colors simultaneously.

In this study, we employ straightforward textual cues to control the eyeglasses style for a more user-friendly and intuitive interface. 
Furthermore, to precisely regulate the spatial configuration of the eyeglasses without introducing undue intricacy, we suggest the implementation of a binary eyeglasses mask as a spatial constraint. 

\section{Method}

Firstly, in Section \ref{framework}, we will provide a comprehensive examination of our proposed framework, including the conditional information encoder, the GlassMapper, and the decoupling strategy. 
Section \ref{loss} will contain a thorough examination of the loss functions implemented during the training process. 
Finally, in Section \ref{scheme}, we will present the two-stage training scheme and its corresponding objective functions.

\subsection{Framework} \label{framework}

\begin{figure*}[!t]
    \centering
    \includegraphics[width=\textwidth]{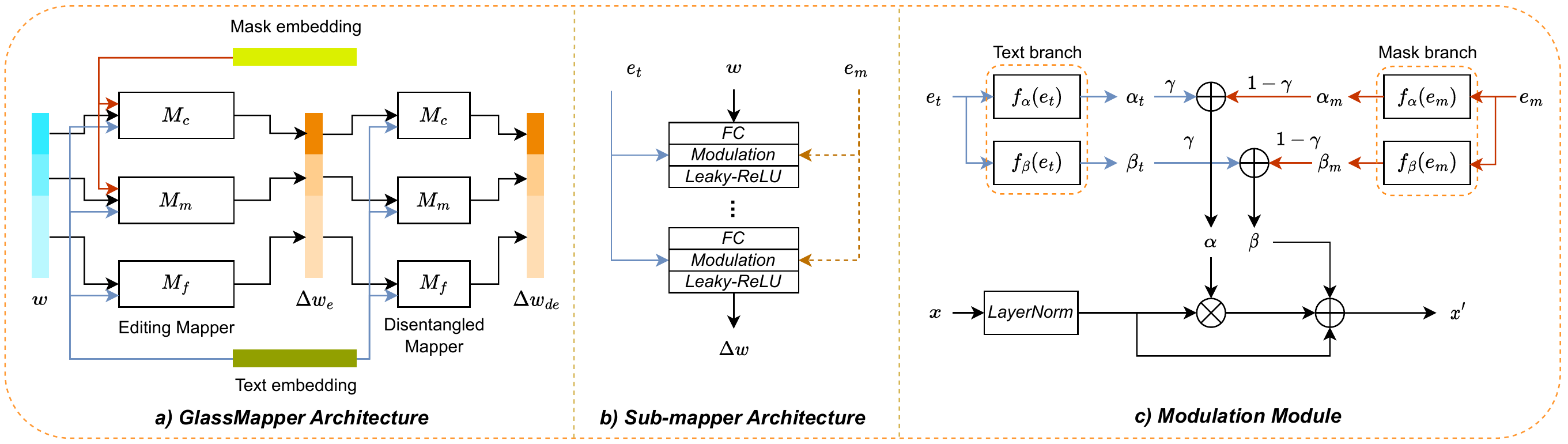}
    \caption{The architecture of GlassMapper. 
    a) GlassMapper is composed of an editing mapper and a disentangled mapper, each of which is divided into three sub-mappers. 
    b) Each sub-mapper consists of 5 blocks, which are stacked with a fully connected layer, a modulation module, and a leaky ReLU activation layer. 
    c) The modulation module is responsible for modulating the input $x \in \mathbb{R}^{512}$ based on the mask condition $e_m \in \mathbb{R}^{512}$ and text condition $e_t \in \mathbb{R}^{512}$. }
    \label{fig: modulation module}
\end{figure*}

Like previous works such as StyleCLIP \cite{styleclip} and HairCLIP \cite{hairclip}, our work utilizes a pre-trained StyleGAN to generate high-quality face images. 
As the $\mathcal{W}+$ space has good semantic decoupling properties, we leverage it to learn a GlassMapper that predicts editing directions to control the shape and style of eyeglasses, given the text prompt and automatically generated mask.

As shown in Fig. \ref{fig: overview framework}, we start by obtaining the source latent code $w_s \in \mathcal{W}+$, the mask embedding $e_m$, and the text embedding $e_t$ through a StyleGAN inversion model, a mask encoder, and a CLIP text encoder, respectively, given a real image $I_{src}$, a eyeglasses mask $m$, and a text description $t$.

The obtained embeddings of different modalities are then sent to the GlassMapper, which consists of an editing mapper and a disentangled mapper. 
The editing mapper predicts an editing direction $\Delta w_e$ based on $e_m$ and $e_t$ while the disentangled mapper predicts a more decoupling editing direction $\Delta w_{de}$ based on $e_t$, given the calculated editing direction $\Delta w_e$.

Finally, the editing latent code $w_{edit} = w_s + \Delta w_e$ and the decoupling editing latent code $w_{de} = w_{edit} + \Delta w_{de}$ are fed into StyleGAN to generate the editing image $I_{edit}$ and the decoupling editing image $I_{de}$.
Specifically, $I_{edit}$ successfully achieves the desired manipulation but unavoidably suffers significant modification in unrelated regions. In contrast, $I_{de}$ inherits the impressive manipulation capability of $I_{edit}$ while preserving irrelevant areas to the greatest extent possible.

\textbf{Conditional Information Encoder}: For text descriptions, a pre-trained CLIP\cite{clip} text encoder is used to extract an expressive text embedding $e_t \in \mathbb{R}^{512}$. 
However, for eyeglasses masks, there is no pre-trained eyeglasses mask encoder available. 
Therefore, a convolutional neural network, similar to that used in MaskGAN\cite{maskgan}, is employed to extract mask conditions. 
The mask encoder consists of $5$ convolutional blocks, each comprising a convolutional layer, an instance normalization layer, and a ReLU activation layer. 
Given an eyeglasses mask, the mask encoder extracts the corresponding mask embedding $e_m \in \mathbb{R}^{512}$. 
These mask and text embeddings are then used to condition the GlassMapper and control the shape and style of eyeglasses in the manipulated image $I_{edit}$ and decoupling editing image $I_{de}$.

\textbf{GlassMapper Architecture}: In Fig. \ref{fig: modulation module}(a), we can see that GlassMapper consists of two distinct mappers: an editing mapper and a disentangled mapper. The editing mapper is specifically designed to manipulate the shape and style of eyeglasses, while the disentangled mapper is responsible for preserving areas that are not relevant to the eyeglasses.

Studies \cite{styleclip, tedigan} have demonstrated that different layers of StyleGAN control different levels of semantic information in the generated image. 
For instance, shallow layers control coarse-grained information like head pose and facial expressions, while deep layers control fine-grained information like color and micro details. 
Therefore, we divide the GlassMapper into three sub-mappers, namely $M_c$, $M_m$, and $M_f$, each controlling different semantic levels in the generated image. 
$M_c$ and $M_m$ are responsible for controlling coarse-grained information like eyeglasses shape, while $M_f$ controls fine-grained information such as eyeglasses color. 
In accordance with this, we split the source latent code $w_s$ into three parts: $w_c$, $w_m$, and $w_f$, which represent different semantic information. 

The editing mapper is purposefully engineered to manipulate the eyeglasses, without being tasked with preserving the irrelevant areas. 
Each sub-mapper of the editing mapper consists of 5 blocks, and each block consists of a fully connected layer, a modulation module, and a leaky relu activation layer, as shown in Fig. \ref{fig: modulation module}(b). 
Since we want to control eyeglasses shape based on the mask, we inject the mask conditions into $M_c$ and $M_m$. 
On the other hand, we use text to control eyeglasses style, which involves fine-grained information like eyeglasses color, so we inject text conditions into all sub-glasses mappers. 
Therefore, the editing mapper can be expressed as:
\begin{equation}
\begin{aligned}
    &M\left(w_s, e_{t}, e_{m}\right) \\ 
    &=\left(M_{c}\left(w_{c}, e_{t}, e_{m}\right), M_{m}\left(w_{m}, e_{t}, e_{m}\right), M_{f}\left(w_{f}, e_{t}\right)\right).
\end{aligned}
\end{equation}

Disentangled mapper is similar to the editing mapper but with two differences: 
1) The number of blocks is set to 2, less than the editing mapper, according to a simpler goal that reduces the modification in the irrelevant areas; 
2) Considering that large modifications in the irrelevant areas are always accompanied by the change of eyeglasses style, we only keep the text branch in the modulation module, ensuring the capability of dealing with different text descriptions. 
Therefore, the disentangled mapper can be expressed as:
\begin{equation}
\begin{aligned}
    &M\left(\Delta w, e_{t}\right) \\ 
    &=\left(M_{c}\left(\Delta w_c, e_{t}\right), M_{m}\left(\Delta w_{m}, e_{t}\right), M_{f}\left(\Delta w_{f}, e_{t}\right)\right).
\end{aligned}
\end{equation}

To incorporate both text conditions and mask conditions into $M_c$ and $M_m$, we modify the original modulation module by introducing a mask branch and a text branch to calculate corresponding scale and bias parameters $\boldsymbol{\alpha_m}$, $\boldsymbol{\beta_m}$, $\boldsymbol{\alpha_t}$, and $\boldsymbol{\beta_t}$, as shown in Fig. \ref{fig: modulation module}(c). 
These parameters are then fused using a weight $\gamma$ to obtain the final scale parameter $\boldsymbol{\alpha}$ and bias parameter $\boldsymbol{\beta}$. 
Finally, the input $\boldsymbol{x}$ is modulated using $\boldsymbol{\alpha}$ and $\boldsymbol{\beta}$ as follows:
\begin{equation}
\begin{aligned}
    &\boldsymbol{\alpha} = \left( 1 - \gamma \right) * \boldsymbol{\alpha_m} + \gamma * \boldsymbol{\alpha_t}, \\
    &\boldsymbol{\beta} = \left( 1 - \gamma \right) * \boldsymbol{\beta_m} + \gamma * \boldsymbol{\beta_t}, \\
    &\boldsymbol{x^{\prime}}=\left(1+\boldsymbol{\alpha} \right) \frac{\boldsymbol{x}-\mu_{x}}{\sigma_{x}}+\boldsymbol{\beta},
\end{aligned}
\end{equation}

{
\setlength{\parindent}{0cm}
where $\mu_x$ and $\sigma_x$ denote the mean and standard deviation of $x$, respectively.
}

\textbf{Simple Decoupling Strategy}: Generating eyeglasses with diverse shapes and styles while preserving irrelevant regions presents a significant challenge. 
Our experimental observations suggest that this objective may compromise the diversity and realism of the generated eyeglasses. 
To address this issue, we propose a simple yet effective decoupling strategy comprising two operations: 
1) We truncate the gradient flow from the disentangled mapper to the editing mapper, thereby liberating the editing mapper from the responsibility of preserving unrelated areas;
2) Based on the outcomes of the editing mapper, we then use the source image as a reference to minimize alterations in unrelated areas while simultaneously preserving the eyeglasses region.
In summary, with the decoupling strategy, the editing mapper mainly focuses on controlling the eyeglasses style while the disentangled mapper mainly focuses on preserving the irrelevant areas, without affecting each other.

\subsection{Loss Function} \label{loss}

In order to meet our objective of controlling the shape and style of eyeglasses in the manipulated image based on mask and text conditions, while preserving the irrelevant areas, we have devised several loss functions. The details are explained below.

{\textbf{Shape Consistency Loss}}: We propose a shape consistency loss that utilizes a face parser network \cite{maskgan} to guide our model in generating eyeglasses that are consistent with the given mask. 
Specifically, we first obtain the segmentation label $S_{src}$ of the source image $I_{src}$ using the face parser network $P$.
Next, we combine $S_{src}$ and the eyeglasses mask $\boldsymbol{m}$  to create the target segmentation label $S_{tar}$, which classifies pixels in the mask region as eyeglasses (excluding the eyes category) while leaving the others unchanged. 
The combination is defined as follows:
\begin{align}
    \left( S_{tar} \right)_{ij} = 
    \begin{cases}
    N_{glasses},&{\text{if}\ m_{ij}=1 \& \left( S_{src} \right)_{ij} \ne N_{eyes} } \\
    \left( S_{src} \right)_{ij},& {\text{otherwise},}
    \end{cases}
\end{align}

{
\setlength{\parindent}{0cm}
where $N_{glasses}$ and $N_{eyes}$ denote the category number of eyeglasses and eyes, respectively.
}

To ensure that the generated eyeglasses have the desired shape conditioned on $\boldsymbol{m}$, we then minimize the cross-entropy loss between $S_{tar}$ and the predicted probability of the manipulated image $I_{edit}$. 
The shape consistency loss is defined as follows:
\begin{gather}
    \mathcal{L}_{sc}=CE \left(P \left( I_{edit} \right), S_{tar}\right),
\end{gather}

{
\setlength{\parindent}{0cm}
where \textit{CE} is cross-entropy loss.
}

\textbf{Classification Loss}: 
To avoid the undesirable phenomenon where GlassMapper relies on shortcuts that inconspicuous and mutilated eyeglasses may match the given mask and achieves a low shape consistency loss, we propose an eyeglasses classification loss. 
GlassMapper leverages this eyeglasses classifier\footnote{\href{https://github.com/apnkv/eyeglasses\_on\_photo}{https://github.com/apnkv/eyeglasses\_on\_photo}} to distinguish between images with and without eyeglasses.
The eyeglasses classification loss is expressed as:
\begin{equation}
    \mathcal{L}_{cls}=Cg\left(I_{edit}\right),
\end{equation}

{
\setlength{\parindent}{0cm}
where \textit{Cg} denotes the eyeglasses classifier. Intuitively, if the classification score decreases, the present probability of eyeglasses will increase, \textit{i.e.}, eyeglasses in the manipulated image will be more fidelity and complete.
}

\textbf{CLIP-NCE Loss}: To perform eyeglasses style manipulation conditioned on the text description, we adopt the CLIP-based Noise Contrastive Estimation (CLIP-NCE) loss\cite{cf-clip}, which encourages the generated eyeglasses to be semantically similar to the input text description. 
It is defined as:
\begin{equation}
    \mathcal{L}_{nce}=-\log \frac{e^{\left(Q \cdot K_{T}^{+} / \tau\right)}+e^{\left(Q \cdot K_{I}^{+} / \tau\right)}}{e^{\left(Q \cdot K_{T}^{+} / \tau\right)}+e^{\left(Q \cdot K_{I}^{+} / \tau\right)}+\sum_{K^{-}}\left(Q \cdot K^{-} / \tau\right)},
\end{equation}

{
\setlength{\parindent}{0cm}
where $\tau$ is the temperature and is set to 1.0. $Q$, $K^+_T$, $K^+_I$ and $K^-$ denote as query, text positive pairs, image positive pairs, and negative pairs, respectively.
}

\textbf{Latent Loss and ID Loss:} Similar to StyleCLIP\cite{styleclip}, we adopt L2 distance to regularize the latent code and constrain the image domain; we also adopt cosine similarity to evaluate the facial feature similarity and preserve the identity information. 
The formulations are defined as follows:
\begin{equation}
    \mathcal{L}_{norm}=\left\|w_{edit}-w_s\right\|_{2},
\end{equation}
\begin{equation}
    \mathcal{L}_{id}=1-\cos \left(R\left(I_{e d i t}\right), R\left(I_{s r c}\right)\right),
\end{equation}

{
\setlength{\parindent}{0cm}
where $R$ denotes ArcFace\cite{arcface} network for face recognition.
}

\textbf{Background Loss}: To preserve the irrelevant areas, we directly compute pixel-wise mean square error between manipulated image $I_{edit}$ and source image $I_{src}$, which is expressed as:
\begin{equation}
    \mathcal{L}_{bg}=\left\|\left(I_{edit}-I_{src}\right) * \left(P_{ng}\left(I_{edit}\right) \cap \left(1-m\right)\right)\right\|_{2},
\end{equation}

{
\setlength{\parindent}{0cm}
where $P_{ng}()$ denotes non-eyeglasses areas mask.
}

\textbf{Disentangled Loss}: When editing the eyeglasses color, we observe a consequent change in cloth color. 
{As the $W+$ space of StyleGAN exhibits strong semantic decoupling properties, we hypothesize that the entanglement issue may arise from the global guidance provided by CLIP, which may bind the color attribute to other objects, rather than eyeglasses.
In other words, CLIP tends to identify an entangled editing direction in which clothing also conforms to color attributes, making it challenging to fully utilize the decoupling properties of $W+$ space.}
Therefore, as the RGB color space is not linearly correlated with human visual perception, we convert all the images from RGB color space to LAB\cite{cielab} color space and define the disentangled loss as:
\begin{equation}
\begin{aligned}
    \mathcal{L}_{disentangle}=&\lambda_{g}\left\|\left(I_{de} -I_{edit}\right) * P_{g}\left(I_{edit}\right)\right\|_{2} \\
    +&\lambda_{c}\left\|\left(I_{de} -I_{src}\right)*P_{c}\left(I_{src}\right)\right\|_{2},
\end{aligned}
\end{equation}

{
\setlength{\parindent}{0cm}
where $P_g()$ denotes eyeglasses area mask and $P_c()$ denotes cloth area mask. 
$\lambda_g$ and $\lambda_{c}$ are set to 4, 5 respectively. 
In this way, we achieve desired eyeglasses style in $I_{de}$ under the supervision of $I_{edit}$, and preserve the irrelevant area under the supervision of $I_{src}$.
}

\begin{figure*}[!t]
    \centering
    \includegraphics[width=0.8\textwidth]{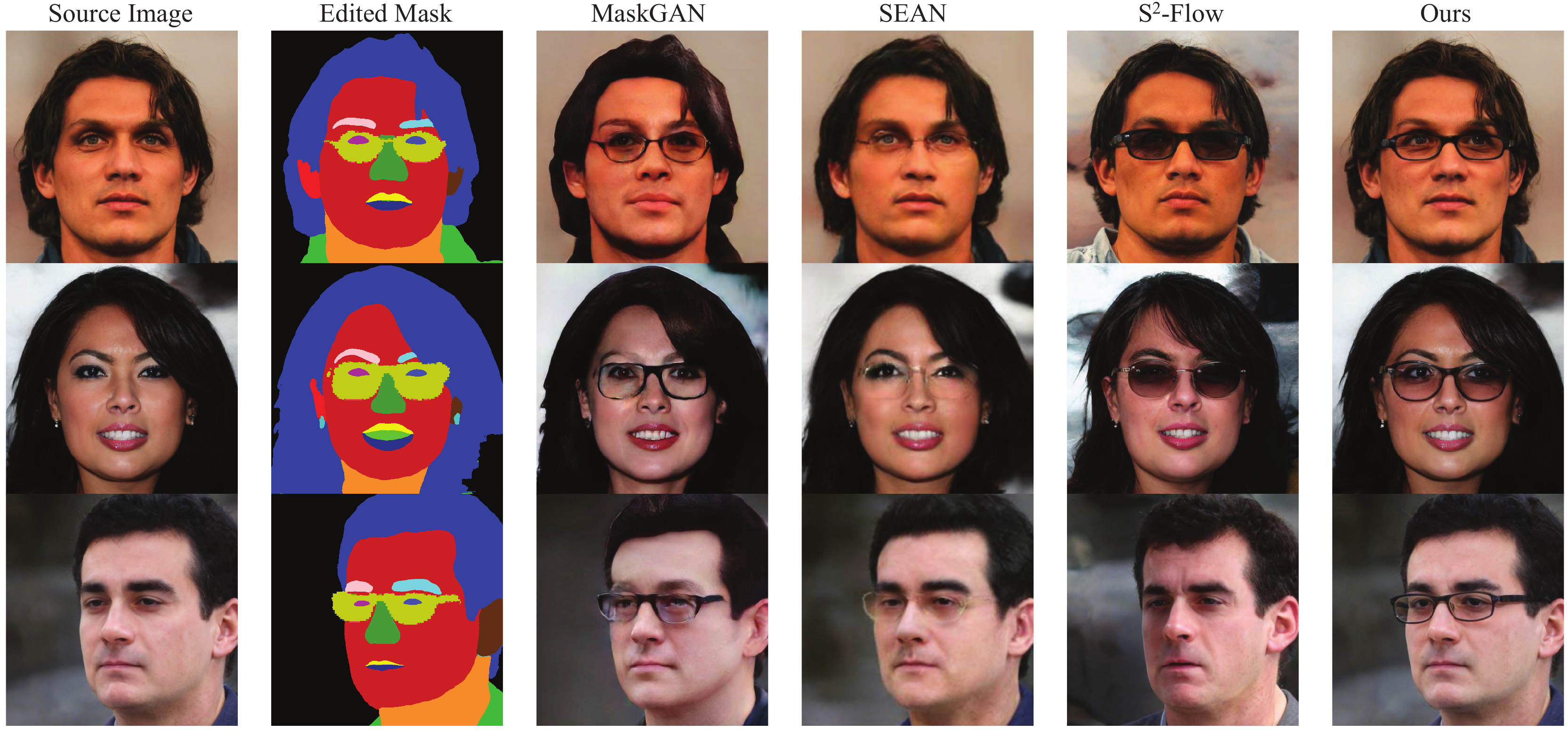}
    \caption{Qualitative comparisons to semantic-based editing methods: {MaskGAN\cite{maskgan}, SEAN\cite{sean}, $S^2$-Flow\cite{S2-flow}}. 
    Our approach successfully generates natural and high-fidelity eyeglasses consistent with the given semantic mask, along with a slight loss of identity information. }
    \label{fig:mask_comparison}
\end{figure*}

\subsection{Training Scheme} \label{scheme}

Intuitively, we can divide the whole task into two sub-tasks, one of which is to control the shape of eyeglasses based on the mask, and the other is to control the style of eyeglasses based on the text.
As we observed that the training convergence speed of each sub-task is distinct, roughly 20 times the difference, it is challenging to train our framework end to end based on the mask and text simultaneously. 

To handle the disparate training convergence rate of different modalities, we adopt a two-stage training approach. 
In the first stage (Stage-I), we train the editing mapper separately for each sub-task to enable preliminary control of the eyeglasses. 
In the second stage (Stage-II), we jointly train the editing mapper to simultaneously achieve all sub-tasks while introducing and learning the disentangled mapper to preserve the irrelevant areas. 
With this approach, we are able to control the eyeglasses shape and style based on the mask and text simultaneously, while also preserving the irrelevant areas.

\textbf{Stage-I}: Considering the distinct training convergence speed of each sub-task (\textit{i.e.}, mask-guided eyeglasses shape manipulation and text-guided eyeglasses style manipulation), we separately train the editing mapper to achieve each sub-task. 
First, we focus on the sub-task of mask-guided eyeglasses manipulation, where we train the editing mapper to control the eyeglasses shape conditioned on the mask. 
Specifically, we set the weight $\gamma$ to 0 and exclude the fine mapper, \textit{i.e.}, only the mask encoder and mask branches in the course mapper and medium mapper are trainable. 
In this way, the objective function is defined as:
\begin{equation}
    \mathcal{L}=\lambda_{sc} \mathcal{L}_{sc}+ \lambda_{\text {cls}} \mathcal{L}_{\text {cls}}+\lambda_{norm} \mathcal{L}_{norm}+\lambda_{id} \mathcal{L}_{id}+\lambda_{\text {bg}} \mathcal{L}_{\text {bg}},
\end{equation}

{
\setlength{\parindent}{0cm}
where $\lambda_{sc}$, $\lambda_{cls}$, $\lambda_{norm}$, $\lambda_{id}$ and $\lambda_{bg}$ are set to 3, 0.03, 0.8, 0.1 and 2, respectively.
}

Later, we focus on the sub-task of text-guided eyeglasses manipulation, where we incorporate text conditions into the editing mapper to modify the eyeglasses style. 
Specifically, we freeze the parameters of the pre-trained mask encoder and mask branches and train all the text branches in the editing mapper. 
Moreover, we also set the weight $\gamma$ to 0.5, which means the mask conditions and text conditions contribute to the editing results equally. 
Finally, the objective function is:
\begin{equation}
    \mathcal{L}=\lambda_{nce} \mathcal{L}_{nce}+
    \lambda_{norm} \mathcal{L}_{norm}+\lambda_{id} \mathcal{L}_{id},
\end{equation}

{
\setlength{\parindent}{0cm}
where $\lambda_{nce}$, $\lambda_{norm}$, $\lambda_{id}$, are set to 0.3, 0.8, and 0.2, respectively.
}

\textbf{Stage-II}: After Stage-I, the editing mapper has preliminary capabilities to achieve each sub-task separately, while performing poorly on the whole task. 
In other words, given a mask and text simultaneously, editing results would be overly modified based on the text, and also inconsistent with the mask. 
Therefore, we jointly train the editing mapper based on the mask and text simultaneously, \textit{i.e.}, all the parameters in the editing mapper are trainable at this stage. 
To preserve the irrelevant areas, we also introduce and learn a disentangled mapper to predict more disentangled editing directions. 
Combined with the decoupling strategy, the editing mapper and the disentangled mapper would perform their duties independently of each other. 
Finally, the total objective function is:
\begin{equation}
    \begin{aligned}
    \mathcal{L} &=\lambda_{nce} \mathcal{L}_{nce}+
    \lambda_{norm} \mathcal{L}_{norm}+\lambda_{id} \mathcal{L}_{id} \\ &+\lambda_{\text{bg}} \mathcal{L}_{\text {bg}}+\lambda_{\text{sc}} \mathcal{L}_{\text{sc}}+
    \lambda_{\text{disentangle}} \mathcal{L}_{\text{disentangle}},
    \end{aligned}
\end{equation}

{
\setlength{\parindent}{0cm}
where $\lambda_{nce}$, $\lambda_{norm}$, $\lambda_{id}$, $\lambda_{bg}$, $\lambda_{sc}$, and $\lambda_{disentangle}$ are set to 0.3, 0.8, 0.2, 5, 4, and 1, respectively.
}

\section{Experiments}

\begin{figure*}[!t]
    \centering
    \includegraphics[width=\textwidth]{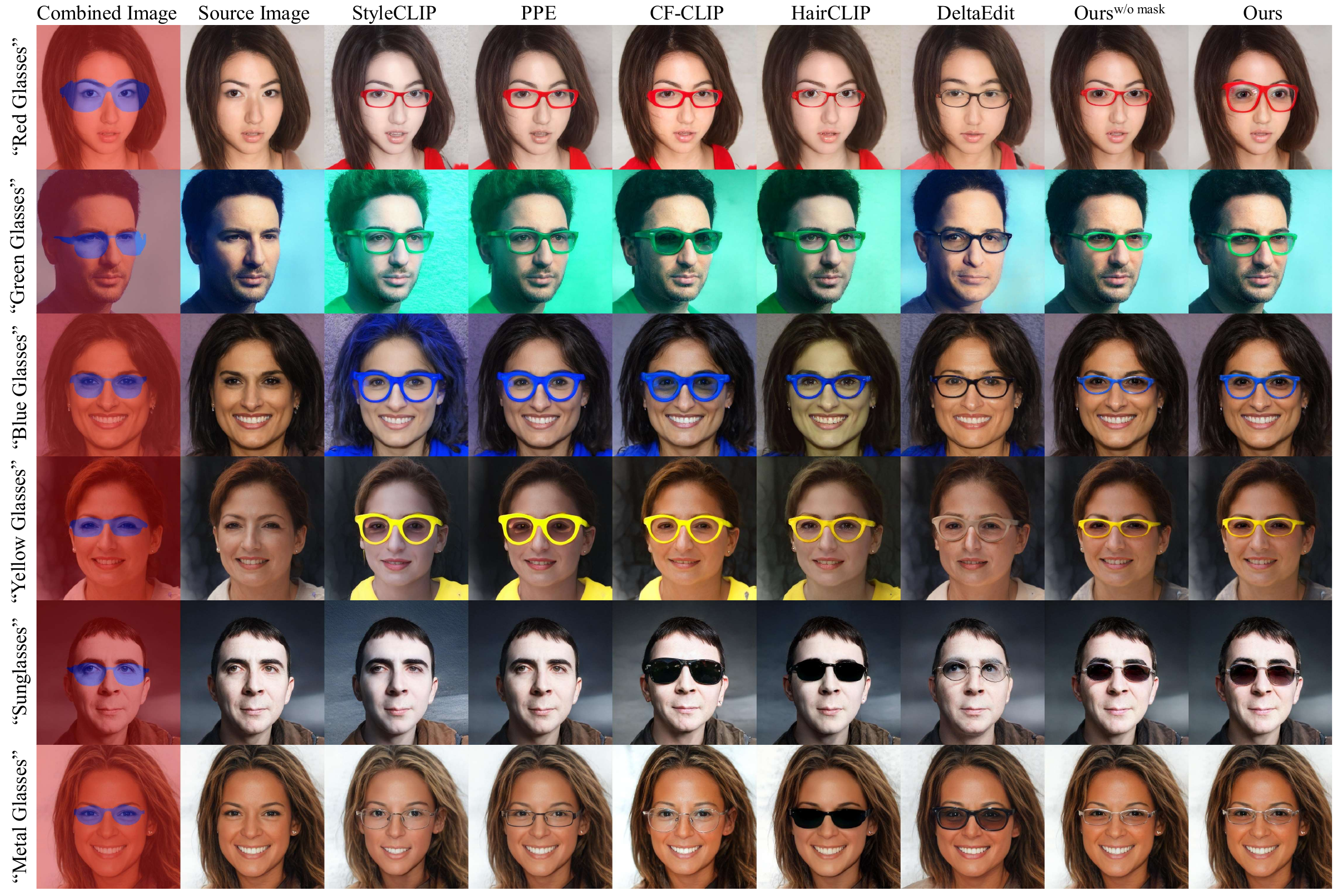}
    \caption{Qualitative comparisons to text-based editing methods: {StyleCLIP\cite{styleclip}, PPE\cite{ppe}, CF-CLIP\cite{cf-clip}, HairCLIP\cite{hairclip}, and DeltaEdit\cite{deltaedit}}. Each row demonstrates the editing results of different methods, and the text conditions are listed on the left side of each row. 
    \textit{Combined Image} denotes the combination of the source image and corresponding eyeglasses mask for better visualization. 
    $Ours^{w/o \ mask}$ denotes we ignore the effect of mask condition.
    Our approach successfully generated proper eyeglasses in all cases along with the least modification in irrelevant areas. 
    It is noteworthy that comparison methods typically produce eyeglasses with identical shapes. 
    In contrast, we can precisely control the shape of the eyeglasses using a binary mask.} 
    \label{fig:qualitative_results}
\end{figure*}

\subsection{Implementation Details}

We use FFHQ dataset\cite{stylegan} for training and CelebA-HQ dateset\cite{celeba-hq} for evaluating. Since we focus on the manipulation of eyeglasses, we split all the datasets based on the presence of eyeglasses. To obtain diverse eyeglasses masks, we utilize a face parser to segment all the images with eyeglasses and get 12,698/1,341 eyeglasses masks in FFHQ/CelebA-HQ datasets. In this way, we construct 12,698 data pairs in FFHQ as the training set and 1,341 data pairs in CelebA-HQ as the test set. \footnote{{Please note that our constructed dataset mainly consists of face images without eyeglasses, binary eyeglass masks, and some textual descriptions of eyeglass styles. Therefore, the training dataset does not contain samples that represent the eyeglass styles discussed in the introduction.}}

We use an 18-layer StyleGAN2\cite{stylegan2} pretrained on FFHQ as our generator and the e4e\cite{e4e} encoder pretrained on FFHQ as our inversion model. 
For text input, we select seven common eyeglasses colors(\textit{i.e.}, red, blue, green, yellow, pink, orange, and purple) and two common eyeglasses styles(\textit{i.e.}, metal glasses and sunglasses). 
As for training, Stage-I is trained for 150,000 iterations, among which we train the mask branches for 145,000 iterations with a base learning rate of 0.005 and train the text branches for 5,000 iterations with a base learning rate of 0.002. 
Stage-II is trained for 20,000 iterations with a base learning rate of 0.001. 

To quantitatively evaluate the performance of our model, we use \textit{Structure Similarity Index Measure} (SSIM)\cite{ssim}, \textit{Peak Signal to Noise Ratio} (PSNR)\cite{psnr}, \textit{Fr\'{e}chet Inception Distances} (FID)\cite{fid}, \textit{Identity Discrepancy Scores} (IDS), \textit{mean of class-wise Intersection over Union} (mIoU), \textit{Pixel Accuracy} (PA), and \textit{CLIP Similarity Score} (CLIPScore) as evaluation metrics. {Intuitively, we used mIoU and PA to evaluate the consistency between the eyeglasses shape and the mask, and CLIPScore to evaluate the alignment between the eyeglasses style and the text description. SSIM, PSNR, IDS, and FID are used to evaluate the preservation of irrelevant regions and identity information. More implementation details can be found in the supplementary material.}

\begin{table}[!t]
\caption{Quantitative comparisons to semantic-based editing methods. 
We achieve better results in most cases, demonstrating a striking editing capability of preserving irrelevant areas and generating high-fidelity eyeglasses consistent with the given semantic mask simultaneously.}
\label{table:quantitative_comparison_mask}
\centering
\setlength{\tabcolsep}{1mm}{
\begin{tabular}{c|cccccc}
\toprule
Method & SSIM($\uparrow$)   & PSNR($\uparrow$)    & IDS($\uparrow$)    & FID($\downarrow$)    & mIoU($\uparrow$)    & PA($\uparrow$) \\
\midrule
MaskGAN\cite{maskgan} & 0.7086 & 20.6624 & 0.2823 & 39.44 & 0.6926 & 0.8985 \\
SEAN\cite{sean} & 0.7365 & 21.3615 & 0.4621 & \textbf{15.30} & 0.7467 & 0.9249  \\
$S^2$-Flow\cite{S2-flow} & 0.6824 & 16.5182 & 0.6057 & 21.81 & 0.5194 & 0.8400 \\
Ours & \textbf{0.8936} & \textbf{27.7014} & \textbf{0.7009} & 27.10 & \textbf{0.7741} & \textbf{0.9584} \\
\bottomrule
\end{tabular}}
\end{table}

\begin{table}[!t]
\caption{Quantitative comparisons to text-based editing methods.
In most cases, we achieve significant improvements, indicating the superior capability of preserving irrelevant areas and identity information.}
\label{table:quantitative_comparison_text}
\centering
\setlength{\tabcolsep}{1.5mm}{
\begin{tabular}{c|ccccc}
\toprule
Method & SSIM($\uparrow$)   & PSNR($\uparrow$)    & IDS($\uparrow$)    & FID($\downarrow$) & CLIPScore($\uparrow$)  \\
\midrule
StyleCLIP\cite{styleclip} & 0.7303 & 17.0802 & 0.5250 & 102.10 & 24.4427 \\
PPE\cite{ppe} & 0.8077 & 20.2526 & 0.5911 & 107.68 & 23.8046 \\
CF-CLIP\cite{cf-clip} & 0.7792 & 19.2501 & 0.5193 & 138.2 & 24.9713 \\ 
HairCLIP\cite{hairclip} & 0.8298 & 23.2001 & 0.6358 & 138.18 & \textbf{25.7187} \\ 
DeltaEdit\cite{deltaedit} & 0.7660 & 18.7950 & 0.4700 & \textbf{48.98} & 20.6276 \\ 
Ours & \textbf{0.8819} & \textbf{26.1910} & \textbf{0.6569} & 94.56 & 24.2708 \\
\bottomrule
\end{tabular}}
\end{table}

\subsection{Qualitative and Quantitative Comparison}

{To achieve diverse and flexible eyeglasses virtual try-on, we propose a method to control eyeglasses shape and style based on a simple binary eyeglasses mask and text description, respectively. Some manipulation methods\cite{maskgan, sean, S2-flow} can edit various facial attributes based on the spatial information provided by the semantic map, while others\cite{styleclip, ppe, cf-clip, deltaedit, hairclip} can edit facial images to align with the given text description. This seems to suggest that existing manipulation methods have the potential to be directly applied to eyeglasses manipulation tasks to achieve arbitrary changes in eyeglasses shape and style. However, eyeglasses manipulation tasks are distinct and challenging, requiring consideration of some special issues, such as maintaining irrelevant areas when changing eyeglasses style, and handling eyeglasses material and reflection effects. Therefore, we chose to compare our method with semantic-based editing methods\cite{maskgan, sean, S2-flow} and text-based editing methods\cite{styleclip, ppe, cf-clip, deltaedit, hairclip} to discuss the unique challenges of eyeglasses manipulation, demonstrate our effectiveness in addressing these challenges, and achieving superior results. }


\textbf{Comparison to Semantic-based Editing Methods}: {The semantic-based editing methods aim to manipulate the source image by utilizing a given semantic map, ultimately aligning the edited image with the semantic map. In the context of our task scenario, we aim to use the semantic map to control the shape and size of eyeglasses in the source image, in order to achieve flexible and precise virtual eyeglasses try-on.}
To evaluate our capability of controlling the eyeglasses shape, we select three semantic-based image editing methods for comparison: MaskGAN\cite{maskgan}, SEAN\cite{sean}, and $S^2$-Flow\cite{S2-flow}. 
All the following comparison experiments are based on their pre-trained models. 

The qualitative results are shown in Fig. \ref{fig:mask_comparison}. 
As illustrated in Fig. \ref{fig:mask_comparison}, our method demonstrates superior performance in terms of consistency in eyeglasses shape and preservation of identity information. 
MaskGAN\cite{maskgan} successfully generates eyeglasses consistent with the edited semantic mask, while the editing results are unnatural with a severe loss of identity. 
On the contrary, SEAN\cite{sean} better preserves the irrelevant areas and identity information while generating low-fidelity eyeglasses. 
Visually, $S^2$-Flow\cite{S2-flow} has a better trade-off between generating high-fidelity eyeglasses and preserving irrelevant areas. 
However, it fails to generate eyeglasses consistent with the given semantic mask in some cases. 
Benefiting from the shape consistency loss, we successfully generate natural and high-fidelity eyeglasses consistent with the given binary mask, along with a slight loss of identity information. 

Table \ref{table:quantitative_comparison_mask} presents the results of our quantitative comparison experiment to evaluate our capability of preserving irrelevant areas and generating high-fidelity eyeglasses, compared to other semantic-based editing methods. 
Compared with other semantic-based editing methods, we achieve a significant improvement in SSIM, PSNR, and IDS metrics, which demonstrates our capability of preserving irrelevant areas and identity information. 
We also get the best results on mIoU and PA metrics, demonstrating the extraordinary capacity to generate high-fidelity eyeglasses consistent with the given semantic mask. 
In terms of FID, we still achieve comparable results, indicating that the distribution of editing results is similar to the source images. 
In total, we demonstrate a striking editing capability of preserving irrelevant areas and generating high-fidelity eyeglasses simultaneously.

\begin{figure}[!t]
    \centering
    \includegraphics[width=0.5\textwidth]{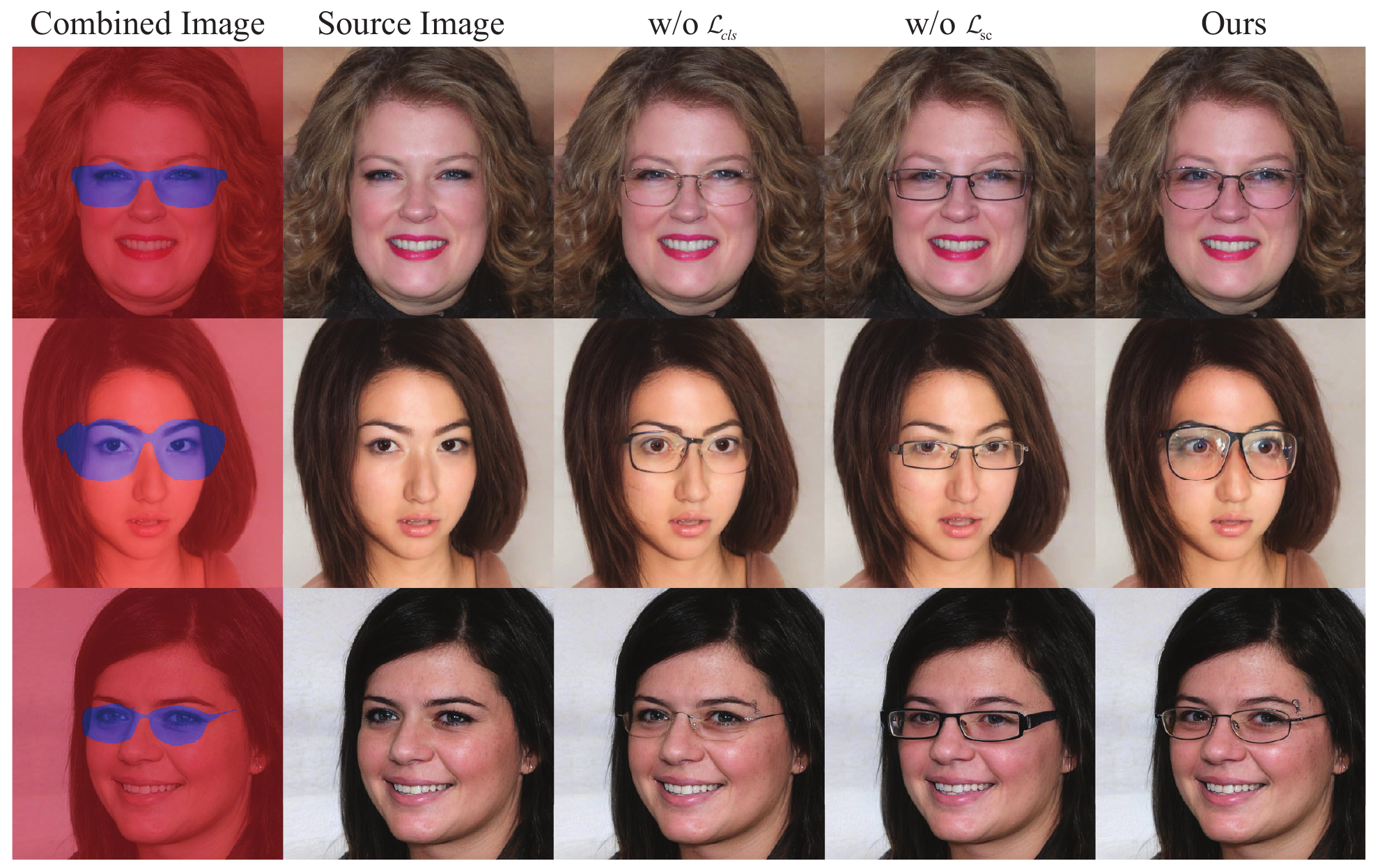}
    \caption{Ablation analysis of $\mathcal{L}_{cls}$ and $\mathcal{L}_{sc}$. 
    \textit{Combined Image} denotes the combination of the source image and corresponding eyeglasses mask for better visualization. 
    By incorporating both the classification loss ($\mathcal{L}_{cls}$) and the shape consistency loss ($\mathcal{L}_{sc}$), we can generate eyeglasses that are more complete and realistic.}
    \label{fig:ablation_cls}
\end{figure}

\textbf{Comparison to Text-based Editing Methods}: {The text-based editing methods aim to manipulate the source image based on a given textual description, ultimately aligning the edited image with the text description. To provide a more convenient and intuitive interactive process, we employ textual descriptions as conditions for modifying different styles of eyeglasses.}
We compare our model to four state-of-the-art text-guided image manipulation methods: StyleCLIP\cite{styleclip}, PPE\cite{ppe}, CF-CLIP\cite{cf-clip}, and DeltaEdit\footnote{In the comparison experiments, we multiply their editing scale since there are no significant alterations in the initial configuration.}\cite{deltaedit}. 
Since these compared methods do not involve mask conditions, we simply focus on the capabilities of modifying the eyeglasses style based on text conditions for a fair comparison, \textit{i.e.}, we do not calculate mIoU and PA during quantitative comparison. 

The qualitative results over CelebA-HQ dataset\cite{celeba-hq} are shown in Fig. \ref{fig:qualitative_results}, our method achieves more natural and realistic results while preserving the irrelevant areas to the greatest extent. 
StyleCLIP\cite{styleclip} successfully achieves most of the eyeglasses styles besides sunglasses, along with a significant modification on irrelevant areas, \textit{e.g.}, hair, skin, and cloth. 
Focusing on disentangling editing, PPE\cite{ppe} reduces changes in most of the irrelevant areas, while still failing in the sunglasses setting. 
CF-CLIP\cite{cf-clip} achieves all the eyeglasses styles we demonstrated, but it may suffer from an excessive modification issue, e.g the lens is also modified in the setting of blue glasses and green glasses. 
{HairCLIP\cite{hairclip} is originally designed for hair style transfer tasks, but with minor modifications, it can also be applied to eyeglass editing. Visually, while Hairclip is capable of achieving various styles of eyeglass editing, it falls short when editing metal eyeglasses, and there are also significant changes in irrelevant areas.}
DeltaEdit \cite{deltaedit}, which is designed to perform arbitrary text-guided image manipulation, fails to produce satisfactory results in any of our experimental settings, even when the editing scale is increased. 
This illustrates the challenge of manipulating eyeglasses based solely on text prompts. 
Meanwhile, all the comparison methods fail to preserve the cloth region when modifying the eyeglasses color. 
Thanks to the disentangled mapper and the simple decoupling strategy, we successfully preserve the cloth region in most cases, regardless of whether we incorporate or ignore the spatial information. 
Moreover, since we control the shape and style of eyeglasses in a single model, we mitigate the excessive modification phenomenon by keeping the balance of each eyeglasses style. 

\begin{figure}[!t]
    \centering    
    \includegraphics[width=0.5\textwidth]{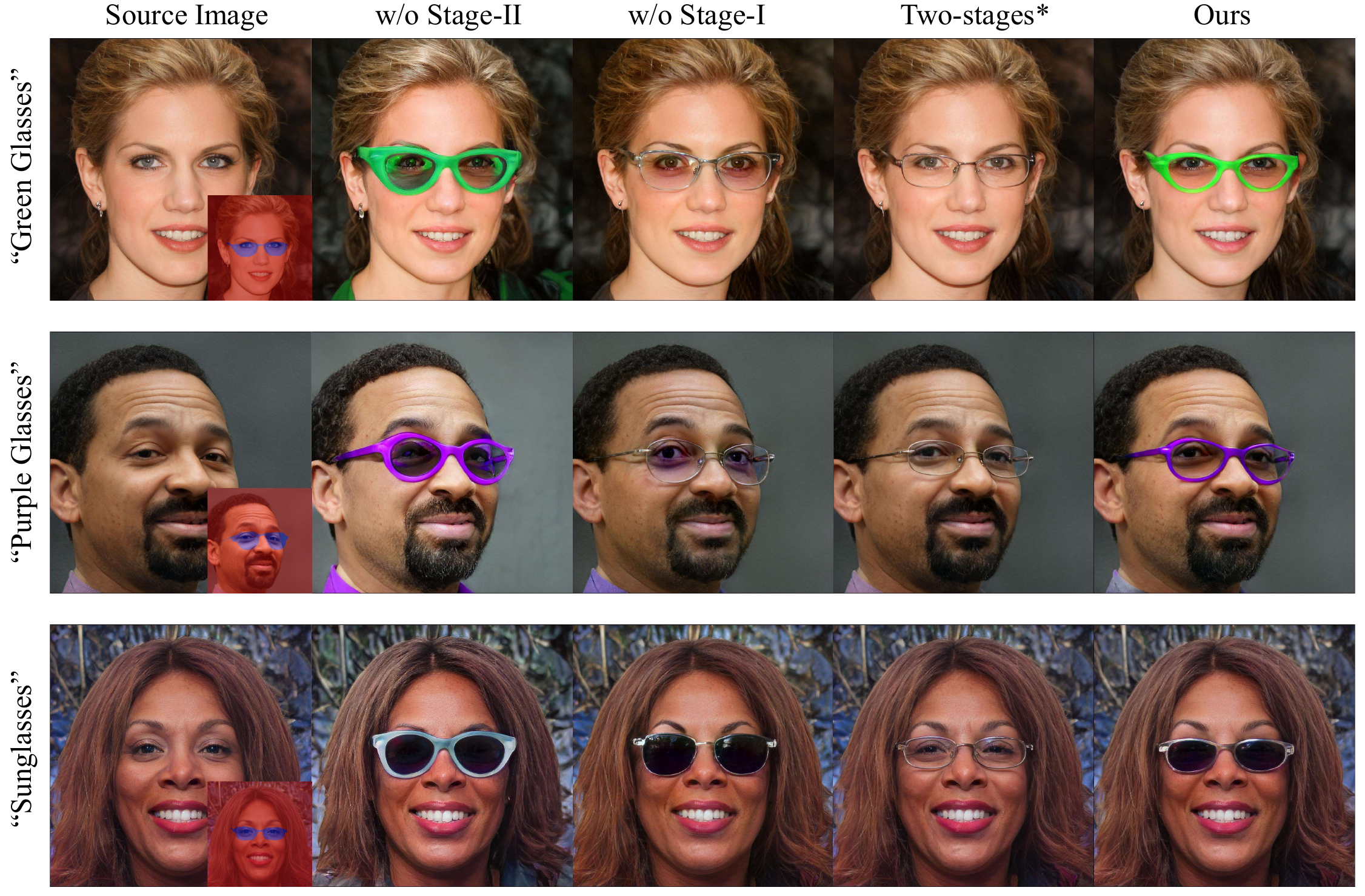}
    \caption{Ablation analysis of the two-stage training scheme. 
    \textit{Two-stage*} denotes we exchange the training orders in Stage-I. 
    Only by using the two-stage training scheme, did we succeed to control the shape and style of the eyeglasses simultaneously.}
    \label{fig:ablation_two_stages}
\end{figure}

\begin{figure}[!t]
    \centering
    \includegraphics[width=0.5\textwidth]{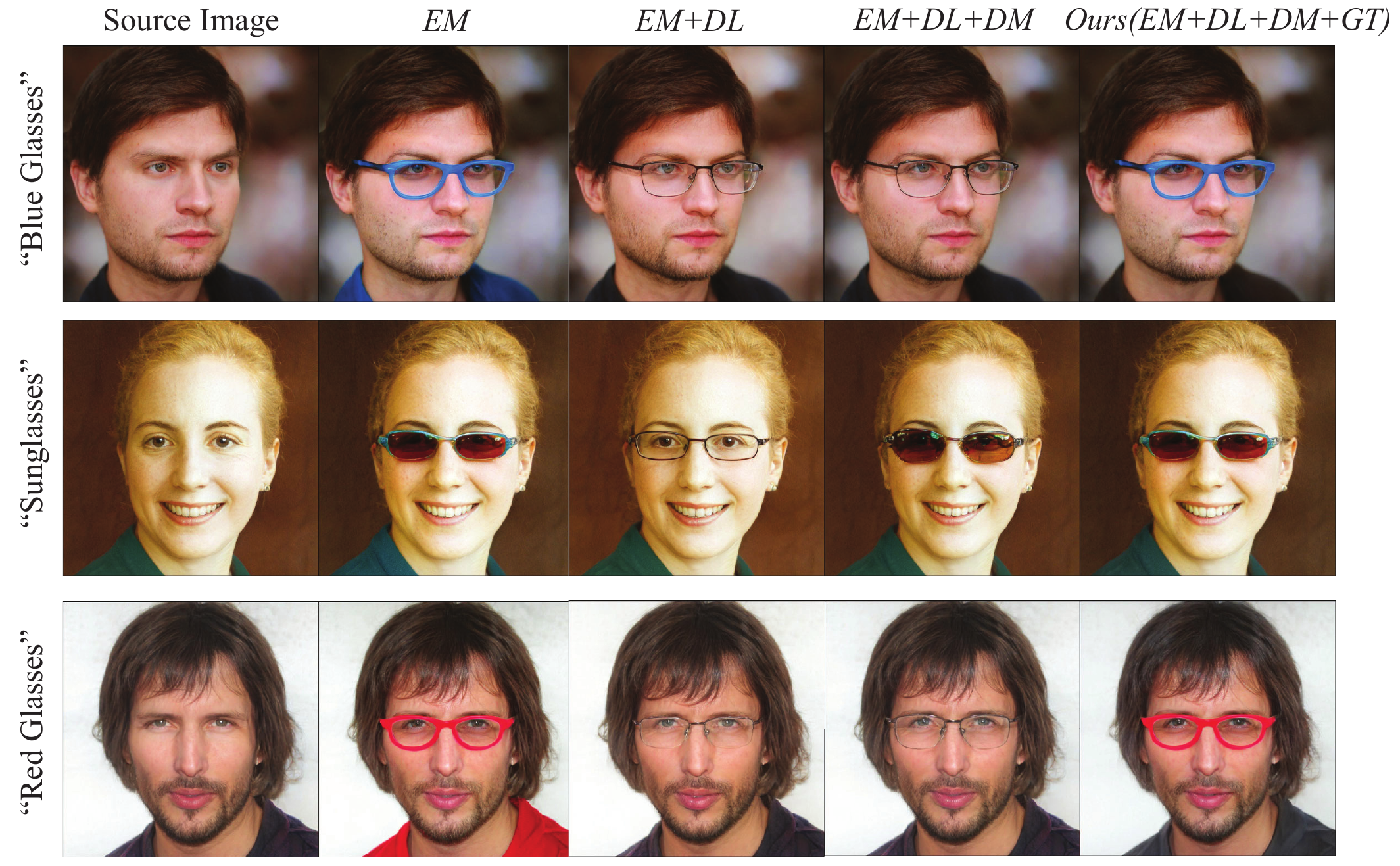}
    \caption{Ablation analysis of the disentangled mapper and decoupling strategy. 
    \textit{EM, DL, DM,} and \textit{GT} denote the editing mapper, disentangling loss, disentangled mapper, and gradient flow truncation, respectively. 
    It indicates that both the disentangled mapper and decoupling strategy is necessary to preserve the irrelevant areas.}
    \label{fig:ablation_disentangle}
\end{figure}

In Table \ref{table:quantitative_comparison_text}, we give the average quantitative comparison results of modifying eyeglasses styles, and more quantitative results are given in the supplementary material.
As is evident, we achieve the best SSIM, PSNR and IDS scores, which are 9.21\%, 29.32\%, and 11.13\% higher than the state of the arts\cite{ppe}, demonstrating our capability of preserving the irrelevant areas and identity information.
All the quantitative results are consistent with our visual comparison, \textit{i.e.}, our approach can not only generate natural and realistic eyeglasses in the editing results but also preserve the irrelevant areas to the greatest extent.
{Despite our approach not achieving the best performance in terms of CLIPScore, it is still comparable to the top results, indicating that we are able to preserve irrelevant areas to a significant extent without sacrificing text control capability.}
We also found that DeltaEdit \cite{deltaedit} achieves the best FID scores, but it fails to produce visually appealing results in all our experimental settings. 
This suggests that FID may not be an appropriate metric for evaluating the manipulation capability of these methods, which is consistent with the findings in \cite{hairclip}.

\textbf{Discussion}: {Given the limited availability of dedicated and open-source methods specifically designed for eyeglass manipulation, we conducted detailed comparative experiments with existing methods\cite{maskgan, sean, S2-flow, ppe, styleclip, cf-clip, hairclip, deltaedit} that address broader issues of face manipulation. It can be observed that the semantic-based editing methods\cite{maskgan, sean, S2-flow} cannot accurately control eyeglass shape conditioned on mask in many cases, while the text-based editing methods\cite{ppe, styleclip, cf-clip, hairclip, deltaedit} fail with some eyeglasses styles and significantly affects irrelevant areas in some cases. This highlights the uniqueness and challenges of the eyeglass manipulation task, where existing methods cannot be easily transferred to this task and require special design. Our method performs well in aligning with different modal information and preserving irrelevant areas, providing a flexible and diverse glasses virtual try-on framework.}

\begin{table*}[!t]
\caption{Quantitative experiments of ablation analysis. Though our method did not achieve the best results in each metric, it was comparable to the optimal results and achieved a balance between aligning with the eyeglasses mask and text description and preserving irrelevant regions.}
\label{table:ablation_study}
\centering
\begin{tabular}{cc|cc|ccc|cccccc}
\toprule
$\mathcal{L}_{cls}$ & $\mathcal{L}_{sc}$ & S-I & S-II & DL & DM & GT & PA($\uparrow$)    & mIoU($\uparrow$) & SSIM($\uparrow$)   & PSNR($\uparrow$)    & IDS($\uparrow$)  & CLIPScore($\uparrow$) \\
\midrule
\checkmark &  &  &  &  &  &  & 0.9404 & 0.6977 & 0.8711 & 25.5300 & 0.6109 & 18.7031  \\
& \checkmark &  &  &  &  &  & 0.9601 & \textbf{0.7835} & \textbf{0.9182} & \textbf{29.0105} & \textbf{0.7358} & 18.7343 \\
\checkmark & \checkmark &  &  &  &  &  & \textbf{0.9617} & 0.7790 & 0.9139 & 28.5235 & 0.6935 & \textbf{19.0625} \\
\midrule
\checkmark & \checkmark & \checkmark &  &  &  &  & 0.9146 & 0.6567 & 0.7746 & 16.7176 & 0.5609 & \textbf{27.5312}   \\
\checkmark & \checkmark &  & \checkmark &  &  &  & \textbf{0.9541} & \textbf{0.7550} & 0.8688 & 25.9677 & \textbf{0.6679} & 20.8125  \\
\checkmark & \checkmark & \checkmark* & \checkmark* &  &  &  & 0.9522 & 0.7455 & \textbf{0.8930} & \textbf{27.3941} & 0.6652 & 18.7503   \\
\checkmark & \checkmark & \checkmark & \checkmark &  &  &  & 0.9538 & 0.7529 & 0.8644 & 23.8862 & 0.6463 & 26.4843  \\
\midrule
\checkmark & \checkmark & \checkmark & \checkmark & \checkmark &  &  & 0.9603 & 0.7776 & 0.9114 & 28.5231 & 0.7073 & 19.0937  \\
\checkmark & \checkmark & \checkmark & \checkmark & \checkmark & \checkmark &  & \textbf{0.9625} & \textbf{0.7849} &\textbf{0.9209} & \textbf{29.3569} & \textbf{0.7282} & 19.1093   \\
\checkmark & \checkmark & \checkmark & \checkmark & \checkmark & \checkmark & \checkmark & 0.9512 & 0.7509 & 0.8736 & 25.5683 & 0.6585 & \textbf{26.3752}  \\
\bottomrule
\end{tabular}
\end{table*}

\subsection{Ablation Analysis}

{To investigate the effects of loss functions, decoupling strategy, and two-stage training scheme on eyeglasses manipulation tasks, we conducted ablation studies that included both quantitative and qualitative analysis. As FID may not be suitable for evaluating manipulation tasks\cite{hairclip}, we ignored the FID results in the quantitative experiments. For more intuitive understanding, we summarized the quantitative results in Table \ref{table:ablation_study}. The specific roles of each component will be analyzed in the following sections, including mask-related loss functions, decoupling strategies, and two-stage training scheme.}

\textbf{Importance of Shape Consistency Loss and Classification Loss}: 
During our two-stage training process, we use several loss functions to control the generated eyeglasses and preserve the irrelevant areas.
We employ widely-used loss functions such as $\mathcal{L}_{nce}$, $\mathcal{L}_{id}$, $\mathcal{L}_{norm}$, and $\mathcal{L}_{bg}$, which have proven their effectiveness in aligning the manipulation results with the text description while keeping irrelevant regions unchanged.
We introduce $\mathcal{L}_{cls}$ and $\mathcal{L}_{sc}$ to better control the shape and alignment of the eyeglasses.

{As shown in the first part of Table \ref{table:ablation_study}, the absence of $\mathcal{L}_{sc}$ leads to poorer mIoU and PA results, indicating that $\mathcal{L}_{sc}$ plays a crucial role in aligning the eyeglasses shape with the given mask. Note that excluding $\mathcal{L}_{cls}$ seems to get better metrics, such as higher mIoU, SSIM, PSNR, and IDS. However, visual examples in Fig. \ref{fig:ablation_cls} demonstrate that dropping $\mathcal{L}_{cls}$ sometimes results in mutilated eyeglasses, indicating an insufficient editing magnitude of the original image in the eyeglasses region. Although it better preserves the irrelevant regions, it also yields results with lower fidelity.
Therefore, both $\mathcal{L}_{cls}$ and $\mathcal{L}_{sc}$ are critical in generating more complete and realistic eyeglasses that are well-aligned with the given mask. 
Dropping either of these loss functions leads to noticeable issues, such as mutilated eyeglasses or misaligned eyeglasses.}


\textbf{Two-Stage Training Scheme}: {To verify the role of our two-stage training scheme, we compare the editing results using different training schemes: 1) We exclude Stage-II, where we do not jointly train our model after the preliminary training of Stage-I; 2) We exclude Stage-I, where we jointly train our model based on the mask and text conditions simultaneously; 3) We adopt another two-stage training scheme, where we exchange the training orders in Stage-I, i.e. we first train our model based on the text conditions and then based on the mask conditions.}

{Visual examples in Fig. \ref{fig:ablation_two_stages} and metrics results in the second part of Table \ref{table:ablation_study} demonstrate the roles of each training stage and the positive impact of adopting a two-stage training strategy on the controllability of eyeglass shape and style.
Firstly, when Stage-II is not utilized, all the editing results are well-aligned with the text descriptions, resulting in a higher CLIPScore. However, the eyeglass shapes differ from the given mask, resulting in lower mIoU and PA results and indicating a loss of control over the eyeglass shape.
Secondly, when Stage-I is not utilized, all the editing results are better consistent with the eyeglass mask, resulting in the best mIoU and PA results. However, we fail to obtain satisfactory results that are well-aligned with the text description, indicating a loss of control over the eyeglass style.
Thirdly, when we exchange the training orders of Stage-I, although we have a better ability to preserve irrelevant regions, it is still difficult to achieve the desired eyeglass style consistently with the text.
As can be seen, it is difficult for all the variants to simultaneously control the shape and style of eyeglasses based on mask and text, and aligning the manipulation results with one modality often results in the loss of control ability of the other modality. 
Nonetheless, our two-stage training scheme can strike a better balance between the control abilities of the two modalities, which leads to a better alignment with the eyeglasses mask and text descriptions.}


\textbf{Impact of Disentangled Mapper and Decoupling Strategy}: To better preserve irrelevant areas, we propose a disentangled mapper and a simple decoupling strategy, \textit{i.e.}, truncate the gradient flow computed by the disentangling loss. 
Therefore, we compare several variants to verify the effect of our disentangled mapper and decoupling strategy, as shown in Fig. \ref{fig:ablation_disentangle} and the third part of Table \ref{table:ablation_study}.

{Solely with the editing mapper (i.e. adopting $\mathcal{L}_{cls}$, $\mathcal{L}_{sc}$ and two-stage training scheme), we successfully achieve satisfactory eyeglasses with desired shape and style, but there present significant modifications in the irrelevant areas, especially the cloth region. 
When including the disentangling loss, we better preserve the irrelevant areas and get higher SSIM, PSNR and IDS results, while failing in generating desired eyeglasses, indicating that the disentangling loss may impair the capability of the editing mapper to generate desired eyeglasses. 
In addition, when incorporating both disentangling loss and disentangled mapper, although the preservation of irrelevant regions is improved, we still encounter the same issue, where the generated eyeglasses significantly deviate from the given text, and the resulting CLIPScore is much lower.
Combining the disentangled mapper and disentangling loss with gradient flow truncation, our approach both obtains diverse eyeglasses styles and preserves the irrelevant areas well, achieving a better balance between the controllability of different modalities (i.e. mask and text) and the preservation of irrelevant regions.}
In other words, by truncating the gradient flow, the editing mapper mainly focuses on controlling the eyeglasses style while the disentangled mapper mainly focuses on preserving the irrelevant areas, without affecting each other.

{Overall, while our approach did not achieve the best performance across all metrics, it demonstrated comparable performance to the optimal results and achieved a balance between the alignment of different modalities and the preservation of irrelevant regions: Firstly, $\mathcal{L}_{cls}$ and $\mathcal{L}_{sc}$ improved the completeness and alignment of the eyeglasses, demonstrating a positive impact on improving the controllability of the mask; Later, two-stage training scheme effectively addressed the mutual interference between the two modalities and achieved a good balance between aligning with the eyeglasses mask and text description; Finally, the disentangled mapper and decoupling strategy maximally preserved the irrelevant regions without sacrificing the control ability of both mask and text. All the experimental results indicate that each component plays a crucial role in eyeglass manipulation tasks and effectively addresses some challenges in this field.}

\begin{figure}[!t]
    \centering
    \includegraphics[width=0.5\textwidth]{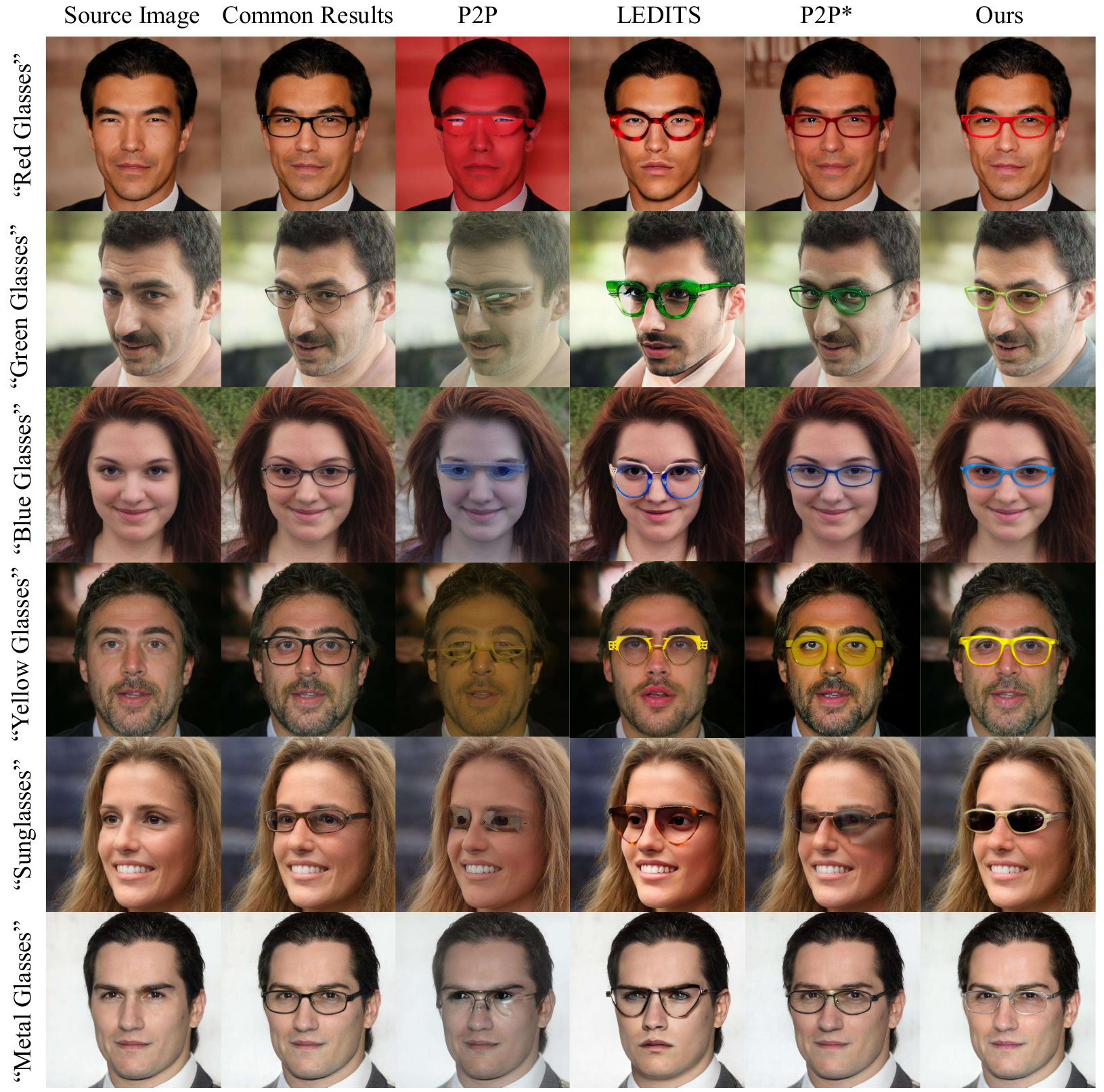}
    \caption{Comparison with the advanced manipulation models. Existing advanced manipulation models\cite{p2p, sega} fall short in fully addressing the specific and challenging task of eyeglass virtual try-on, while our method achieves superior results.}
    \label{fig:diffusion_comparison}
\end{figure}

\subsection{Exploring the Potential of Advanced Models in Eyeglasses Manipulation}

{With the rapid advancements in diffusion models, large text-to-image models\cite{ldm, dalle2, imagen} have demonstrated remarkable generation capabilities, producing stunning and high-quality images. 
Building upon these impressive generation abilities, researchers have explored their potential in the field of image editing, aiming to achieve more versatile and comprehensive image manipulation. 
Notable studies in this area include Prompt-to-Prompt\cite{p2p} and SEGA\cite{sega}. 
Specifically, Prompt-to-Prompt utilizes the semantic relationship in the cross-attention layers between pixels and tokens to implement three editing modes: word swap for local editing, new phrase addition for global editing, and attention re-weighting for extent controlling. 
SEGA demonstrates how text interacts with the diffusion process and propose an approach to discover semantic directions in a single forward pass, allowing for local and global editing in composition and style.
Given the notable progress in advanced manipulation models, we were motivated to investigate their potential in the specific task of eyeglasses virtual try-on.}

{In order to achieve real image editing, we utilized Prompt-to-Prompt (combined with null-text inversion\cite{null-text}) and LEDITS\cite{ledits} (an expanded version of SEGA) in our experimental setting for eyeglasses manipulation.
As illustrated in Fig. \ref{fig:diffusion_comparison}, when eyeglasses-related phrases were directly added in the text, Prompt-to-Prompt \cite{p2p} always resulted in global editing and it was difficult to achieve the desired eyeglass styles. Compared to Prompt-to-Prompt, LEDITS \cite{ledits} achieved better results; however, the generated eyeglasses were still unrealistic and resulted in a loss of identity information. To further explore the local editing ability of Prompt-to-Prompt, we used a simpler experimental setting where we edited the existing eyeglasses in the source images, rather than directly generating eyeglasses with a specified style\footnote{Common results with regular eyeglasses were generated by our approach, where the differences compared to the source images may be visually negligible.}. As can be seen, under such a simpler experimental setting, Prompt-to-Prompt can maintain irrelevant areas and identity information well, but the edited eyeglasses were not entirely satisfactory, such as sunglasses, metal eyeglasses, and yellow eyeglasses. In contrast, our method can not only directly generate diverse and high-quality eyeglasses but also maintain irrelevant areas and identity information to a great extent.}

{Undoubtedly, advanced manipulation models\cite{p2p, sega, inversionedit} have extraordinary editing capabilities and can perform various types of editing on different types of images. However, for editing tasks that involve adding small objects, especially for eyeglass virtual try-on, existing methods are still far from satisfactory. In other words, eyeglass virtual try-on is a specific and challenging task that cannot be fully addressed by these advanced manipulation models. Although our method are not as versatile as these models, it can achieve better results for the specific task of eyeglass virtual try-on.}

\begin{table}[!t]
\caption{Inference time analysis. Our approach can achieve flexible and diverse eyeglass virtual try-on under the control of both mask and text inputs without incurring significant inference time costs.}
\label{table:infer_time}
\centering
\begin{tabular}{c|ccccc}
\toprule
 & \makecell[c]{mask \\ control} & \makecell[c]{text \\ control} & \makecell[c]{retraining \\ per style} & \makecell[c]{inference \\ time (sec)} \\
\midrule
MaskGAN\cite{maskgan} & \checkmark &  & no & \textbf{0.0199\boldmath$\pm$0.0796} \\
SEAN\cite{sean} & \checkmark &  & no & 0.3602$\pm$0.0530  \\
$S^2$-Flow\cite{S2-flow} & \checkmark &  & no & 0.5378$\pm$0.1058 \\
PPE\cite{ppe} &  & \checkmark & yes & 0.0952$\pm$0.0278 \\
StyleCLIP\cite{styleclip} &  & \checkmark & yes &  0.0976$\pm$0.0210 \\
CF-CLIP\cite{cf-clip} &  & \checkmark & yes & 0.1273$\pm$0.0272 \\
HairCLIP\cite{hairclip} &  & \checkmark & no & 0.1171$\pm$0.0073 \\
DeltaEdit\cite{deltaedit} &  & \checkmark & no & 0.0994$\pm$0.0006 \\
Ours & \checkmark & \checkmark & no & 0.1168$\pm$0.0069  \\
\bottomrule
\end{tabular}
\end{table}

\subsection{Inference Time Analysis}

{In this section, we conducted a comprehensive performance evaluation of our proposed approach and most comparison methods, specifically focusing on the inference time required for editing a single image. Specifically, we utilized an RTX 1080 GPU and measured the overall duration of the model, from input to output, for each test sample. We then computed the average inference time based on these measurements.}

{As demonstrated in Table \ref{table:infer_time}, while our method may not be the fastest in terms of inference speed, it still achieves excellent results that are comparable to other methods\cite{hairclip, styleclip, ppe, cf-clip, deltaedit}. It is important to note that our approach can manipulate both the eyeglass shape using the mask input and the eyeglass style using the text input, without necessitating retraining a new model for each style, showcasing the efficiency of our approach. In summary, following a single training session, our approach enables flexible and diverse eyeglass try-on, controlled by both mask and text inputs, without incurring significant inference time costs.}

\section{Conclusion}
Focusing on eyeglasses virtual try-on, we propose a Text-guided Eyeglasses Manipulation method to control the eyeglasses shape and style based on the mask and text, which is intuitive and effective. 
Thanks to the new modulation module and two-stage training scheme, the mask conditions and text conditions control the eyeglasses simultaneously in a decoupling way, which is a more flexible and controllable interaction way. 
Furthermore, by utilizing the simple but effective decoupling strategy, we preserve the irrelevant areas to a great extent, resulting in better local editing. 
Quantitative and Qualitative Comparison and ablation analysis demonstrate the superiority of our approach to achieve diverse and realistic eyeglasses and preserve the irrelevant areas in the editing results.

Future research may focus on streamlining the training process, which currently necessitates separate training for each modality followed by joint training. 
Another potential avenue for improvement is the generalization to more challenging cases, such as highly regular eyeglasses shapes or unusual eyeglasses styles. 
The collection of additional extreme eyeglasses masks and unusual eyeglasses prompts may provide a viable solution.


\bibliographystyle{IEEEtran}
\bibliography{ref_whole}
\newpage

\vfill

\end{document}